\documentclass[11pt]{article}

\usepackage[final]{acl}

\usepackage{times}
\usepackage{latexsym}

\usepackage[T1]{fontenc}

\usepackage[utf8]{inputenc}

\usepackage{microtype}

\usepackage{inconsolata}

\usepackage{graphicx}
\usepackage{algorithm}
\usepackage{algpseudocode}
\usepackage{url}
\usepackage{amssymb}
\usepackage[]{amsmath}
\usepackage{mathtools}
\usepackage{multirow}
\usepackage{booktabs}
\usepackage{caption}
\usepackage{cuted} 
\usepackage[table]{xcolor}

\title{\textsc{TreePS-RAG}: Tree-based Process Supervision for \\ Reinforcement Learning in Agentic RAG}

\author{Tianhua Zhang$^{\heartsuit}$\thanks{$\;\;$Equal contribution.}$\,$, Kun Li$^{\heartsuit*}$, Junan Li$^{\heartsuit*}$, Yunxiang Li$^{\heartsuit}$, \\ \bf
Hongyin Luo$^{\diamondsuit}$, Xixin Wu$^{\heartsuit}$, James Glass$^{\diamondsuit}$, Helen Meng$^{\heartsuit}$ \\
$^\heartsuit$The Chinese University of Hong Kong, Hong Kong SAR, China \\
$^\diamondsuit$Massachusetts Institute of Technology, Cambridge MA, USA \\
\texttt{\{thzhang, li.kun\}@link.cuhk.edu.hk, jli@se.cuhk.edu.hk}
}

\begin{document}
\maketitle
\begin{abstract}
Agentic retrieval-augmented generation (RAG) formulates question answering as a multi-step interaction between reasoning and information retrieval, and has recently been advanced by reinforcement learning (RL) with outcome-based supervision. While effective, relying solely on sparse final rewards limits step-wise credit assignment and provides weak guidance for intermediate reasoning and actions. 
Recent efforts explore process-level supervision, but typically depend on offline constructed training data, which risks distribution shift, or require costly intermediate annotations. 
We present \textbf{\textsc{TreePS-RAG}}, an online, tree-based RL framework for agentic RAG that enables step-wise credit assignment while retaining standard outcome-only rewards. Our key insight is to model agentic RAG reasoning as a rollout tree, where each reasoning step naturally maps to a node. This tree structure allows step utility to be estimated via Monte Carlo estimation over its descendant outcomes, yielding fine-grained process advantages without requiring intermediate labels. To make this paradigm practical, we introduce an efficient online tree construction strategy that preserves exploration diversity under a constrained computational budget.
With a rollout cost comparable to strong baselines like Search-R1, experiments on seven multi-hop and general QA benchmarks across multiple model scales show that \textsc{TreePS-RAG} consistently and significantly outperforms both outcome-supervised and leading process-supervised RL methods.
\end{abstract}

\section{Introduction}
Recent years have witnessed substantial progress in large-scale generative language models, enabling strong performance across many NLP tasks from natural-language prompts \cite{brown2020languagemodelsfewshotlearners,ouyang2022traininglanguagemodelsfollow}. However, reliability remains a key bottleneck: models prone to hallucinate and often lack verifiable provenance for factual claims, which is especially problematic in knowledge-intensive settings such as open-domain question answering \cite{Huang_2025, zhang2025sirenssongaiocean}. Retrieval-augmented generation (RAG) mitigates these issues by grounding generation on externally retrieved evidence to improve factual coverage \cite{lewis2021retrievalaugmentedgenerationknowledgeintensivenlp,gao2024retrievalaugmentedgenerationlargelanguage,fan2024surveyragmeetingllms,li-etal-2025-generate}.

While conventional RAG pipelines often follow single-turn retrieval or relatively static workflows, they struggle on complex questions that require iterative reasoning or dynamic retrieval intent. This has motivated \emph{agentic RAG} \cite{singh2025agenticretrievalaugmentedgenerationsurvey}, where Large Language Models (LLMs) interleave reasoning and information-seeking actions, often instantiated through ReAct-style loops \cite{yao2023reactsynergizingreasoningacting}. Beyond prompt-only approaches \cite{trivedi2023interleavingretrievalchainofthoughtreasoning,madaan2023selfrefineiterativerefinementselffeedback,gou2024criticlargelanguagemodels}, recent work increasingly treats agentic RAG as a trainable decision-making problem \cite{asai2023selfraglearningretrievegenerate,li-etal-2025-survey}. In particular, outcome-supervised reinforcement learning has emerged as a scalable paradigm for eliciting reasoning and planning behaviors \cite{deepseekai2025deepseekr1incentivizingreasoningcapability}, and has been adapted to train search-enabled agents that learn how to retrieve information \cite{chen2025researchlearningreasonsearch,song2025r1searcherincentivizingsearchcapability}. Within this line, Search-R1 serves as a representative approach: it treats retriever as a tool in the RL environment and optimizes multi-turn search-generation interactions using a trajectory-level reward defined by final answer correctness \cite{jin2025searchr1trainingllmsreason}.

Despite the promise of agentic RAG, a prevalent training choice is to execute decisions step by step while optimizing with an outcome-only signal computed at the end of the trajectory (e.g., final answer correctness)\cite{song2025r1searcherincentivizingsearchcapability, jin2025searchr1trainingllmsreason, luo2025infoflowreinforcingsearchagent}. Such delayed and sparse supervision exacerbates credit assignment: intermediate search and reasoning decisions are updated as if they were equally responsible for the terminal outcome, even though only a subset of steps are truly decision-critical. This raises our central question: \emph{can process supervision improve the learning 
of search-enabled agents beyond outcome-only optimization?} A natural approach is to introduce process supervision via step-wise rewards; however, as summarized in \citet{tran2025exploitingtreestructurecredit}, recent step-reward formulations often face two practical limitations: (i) they require explicit step-level annotations to supervise intermediate decisions \cite{zheng-etal-2025-stepsearch}, and (ii) the assignment of process rewards is frequently carried out offline (post hoc over collected trajectories), rather than being integrated into online RL rollouts for interactive agents \cite{zhang2025process}.

To address these limitations, we introduce \textsc{TreePS-RAG} (\textbf{Tree}-based \textbf{P}rocess-\textbf{S}upervised \textbf{RAG}), a tree-structured reinforcement learning approach for agentic retrieval that provides process-level supervision without intermediate annotations or auxiliary reward models. 
The key idea is to  model the agentic RAG rollouts as a tree formulation,
where each node corresponds to one search step. During training, \textsc{TreePS-RAG} assigns node-level supervision on the fly via Monte Carlo estimation over descendant leaves: it samples multiple continuations from a node, scores terminal outcomes by exact match (EM), and propagates the aggregated outcome signal back as the node’s value for advantage calculation. This yields denser step-wise supervision while preserving standard outcome-based policy optimization.
To keep online tree expansion computationally tractable, we further introduce similarity-based pruning over retrieved evidence, pruning nodes whose retrieved document sets are highly overlapping to control the branching factor while maintaining diverse paths. We evaluate \textsc{TreePS-RAG} on seven QA benchmarks under four backbone LLMs, and observe consistent improvements over competitive baselines across datasets and models, indicating robust benefits from process supervision for agentic RAG.

In summary, our contributions are threefold: (i) we introduce an online tree-structured rollout for agentic RAG that enables effective exploration at a cost comparable to outcome-based RL;
(ii) we derive online, annotation-free process supervision from descendant terminal outcomes without auxiliary reward/value models; and (iii) on seven QA benchmarks, our method consistently outperforms competitive baseline training approaches.

\section{Methodology}
\begin{figure*}[ht]
\centering
\includegraphics[width=1\textwidth]{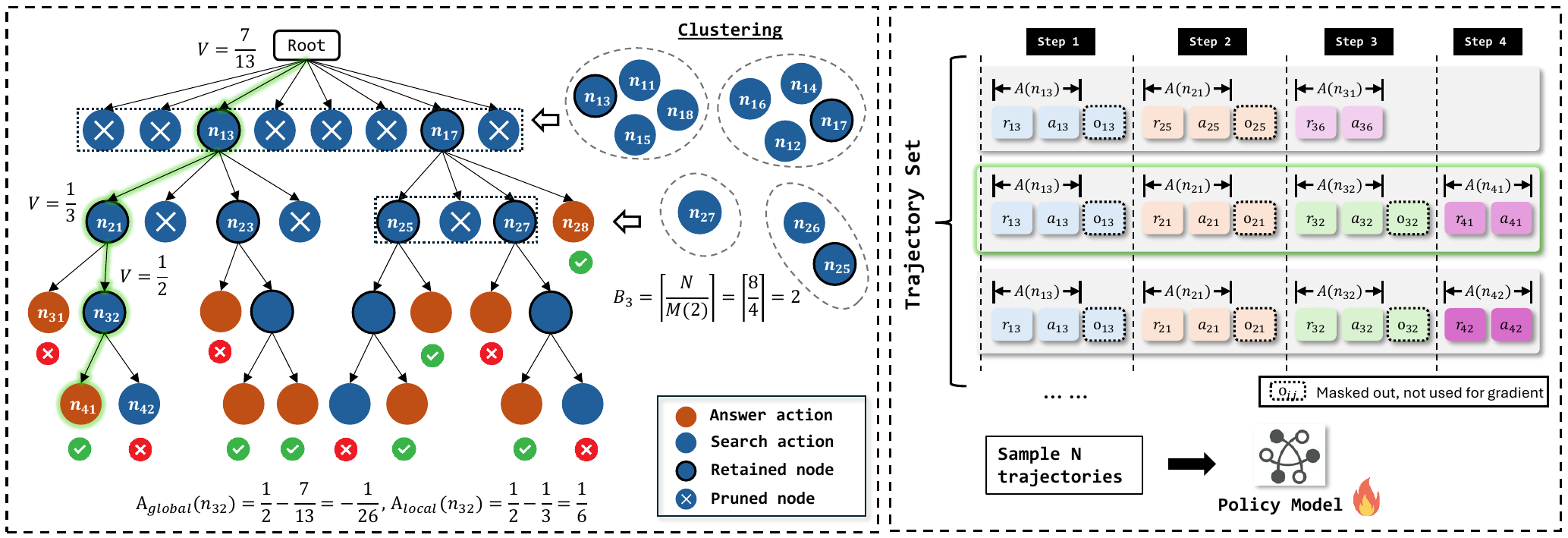}
\caption{Overview of \textsc{TreePS-RAG}. 
\textbf{Left:} Online tree construction and process supervision with $N=8$, $D=4$,  $N_{\text{retain}}=2$. 
Retained parent nodes are expanded with $B_d$ continuations. 
Similarity-based clustering is applied among sibling \textit{search} children
to prune redundant nodes (crossed out) and retain $\hat{N}_{\text{retain}}$ diverse candidates (solid circles).
Node values ($V$) are estimated via Monte Carlo averaging over descendant leaf outcomes to compute process advantages.
\textbf{Right:} Tree-to-trajectory transformation for policy optimization. 
Root-to-leaf paths are collected into a set (one example highlighted in green). Process advantages are uniformly assigned to
all model-generated tokens within a step $(r_i, a_i)$, while observation tokens $(o_i)$ are masked out. 
For fair comparison, only $N$ trajectories is randomly sampled during training to maintain comparable to standard outcome-based RL methods. 
}

\label{fig: method-example}
\end{figure*}

We present \textsc{TreePS-RAG}, an online tree-based reinforcement learning framework for agentic RAG that enables process-level credit assignment without requiring intermediate annotations, auxiliary reward or value models. \textsc{TreePS-RAG} improves step-wise supervision through tree exploration, while retaining the standard outcome-based policy optimization. We first formalize the conceptual approach, modeling agentic RAG inference as a rollout tree
(\S \ref{method: tree-modeling}), then introduce an efficient online tree construction strategy (\S \ref{method: online-tree-construction}), and finally detail how to derive process-level supervision (\S \ref{method: process-supervision}). Figure \ref{fig: method-example} and Algorithm \ref{alg:main} illustrate the overall workflow.

\subsection{Tree Modeling of Agentic RAG}
\label{method: tree-modeling}
Agentic RAG under the ReAct \cite{yao2023reactsynergizingreasoningacting} paradigm solves a given question $q$ in a sequential, step-wise manner. A language model agent $\pi_{\theta}$ alternates between reasoning, information-seeking via an external retriever, and answer generation. At each step $i$, the agent conditions on the current state $s_i=[x,(r_1,a_1,o_1),...(r_{i-1},a_{i-1},o_{i-1})]$, which consists of the prompted input containing the original question (Figure \ref{fig: input-prompt}) together with all accumulated interaction history, to produce the next reasoning segment $r_i$ along with an action $a_i \in \{\textit{search}, \textit{answer}\}$.
When a \textit{search} action is generated, the agent issues the query to the retriever and receives an observation $o_i$, consisting of the top-$\text{K}$ retrieved passages. A trajectory $y=[(r_1,a_1,o_1), (r_2, a_2, o_2), ...]$ terminates when the agent produces an \textit{answer} action or reaches a predefined maximum step limit. Upon termination, the agent receives a scalar outcome reward $r(y)$, reflecting the correctness of the final answer. The goal is to optimize the policy $\pi_{\theta}$ by maximizing the expected cumulative reward.

\paragraph{Tree Representation of Rollouts} We model the agentic RAG rollout process as a tree rooted at the input prompt $x$. Each agentic step $i$, comprising the reasoning-action pair together with retrieved observation when applicable, i.e., $(r_i,a_i,o_i)$, is represented as a node $n_i$ in the tree. Nodes associated with \textit{search} actions allow subsequent exploration, while each node with \textit{answer} action forms a leaf. In this way, a complete trajectory $y$, composed of a sequence of successive actions, corresponds to a root-to-leaf path in the tree. Under this formulation, step-wise reasoning process naturally maps to parent-child relationships. Expanding a node amounts to sampling continuations from the policy:
\begin{equation}
\label{eq:policy-sampling}
\scalebox{0.9}{$
\begin{aligned}
(r_i, a_i)
&\sim
\pi_\theta\bigl(
\cdot \mid x,
(r_1,a_1,o_1), \ldots,
(r_{i-1},a_{i-1},o_{i-1})
\bigr)
\end{aligned}
$}
\end{equation}

\paragraph{Why Tree Modeling?}
This tree-structured formulation is particularly well-suited for agentic RAG under outcome-only supervision.
First, the sequential nature of agentic RAG admits a direct mapping of reasoning steps separated by actions to tree nodes. More crucially, in the absence of step-level annotations, the quality of an intermediate action cannot be directly observed. A tree provides a principled mechanism to address this limitation: by treating the path to a given step node as a shared prefix and exploring multiple continuations from that point, the utility of the corresponding step can be estimated empirically from the outcomes of its descendant trajectories. This perspective closely aligns with Monte Carlo-style estimation, where the value of a state is inferred from the returns of sampled rollouts originating from it. Under this view, the tree is not merely a data structure, but an active computational tool that enables exploration and step-wise credit assignment for agentic RAG.


\subsection{Practical Online Tree Construction}
\label{method: online-tree-construction}
While the tree formulation introduced in \S \ref{method: tree-modeling} offers a principled view of agentic RAG rollouts, naively expanding all nodes in an exhaustive way quickly becomes computationally intractable due to the exponential growth of the tree.
To make tree-based exploration feasible under realistic training budgets, we propose an efficient online tree construction strategy that explicitly controls the branching and selectively retains informative continuations. This preserves the advantages of tree modeling, i.e., shared prefixes and multi-branch exploration, while maintaining a rollout cost comparable to conventional outcome-based RL methods such as Search-R1 \cite{jin2025searchr1trainingllmsreason}.  
Specifically, our tree construction is governed by three core parameters: 
(1) a target rollout budget $N$, chosen to match the token cost of rollout in those methods that conventionally sample N trajectories in parallel, (2) a maximum depth $D$, limiting the number of agentic steps, and (3) a local retention budget $N_{retain}$, which controls how many non-leaf child nodes are retained for each parent.

\paragraph{Initialization} The tree is initialized with the input prompt as the root $n_{\texttt{root}}$. Let $M(d)$ denote the set of retained non-leaf nodes at depth $d$, which serves as parents for expansion at the next step. Initially, $M(0)=\{n_{\texttt{root}}\}$. The tree is expanded in a layer-wise manner, and each expansion from depth $d-1$ to $d$ follows a generate-then-prune paradigm.

\paragraph{Child Node Generation} At depth $d$, for each parent node $n_p \in M(d-1)$, the language model generates $B_d$ new children with Eq. (\ref{eq:policy-sampling}) in parallel. This forms a local children set $C(n_p)=\{n_p^1, ..., n_p^{B_d}\}$, where each child corresponds to a sampled continuation step. The branching factor $B_d$ is dynamically computed as: 
\begin{equation}
    B_d=\left\lceil \frac{N}{\lvert M(d-1) \rvert} \right\rceil 
\label{eq: branch-factor}
\end{equation}
This depth-wise allocation ensures that the total number of sampled nodes at each layer remains approximately $N$, i.e., $\forall d \in [1, D], \lvert M(d-1) \rvert \times B_d \approx N$.\footnote{Due to the existence of \textit{answer} actions and the ceiling operation, this number is not strictly fixed to $N$ but fluctuates around it, ensuring the overall computational cost remains comparable to sampling $N$ independent trajectories.} As a result, the overall rollout cost remains comparable to sampling $N$ independent trajectories in conventional rollout mechanisms.

\paragraph{Child Node Pruning} Among the children of a given parent, each node associated with the \textit{answer} action immediately forms a leaf and defines a complete trajectory corresponding to the root-to-itself path. For the remaining \textit{search}  children, denoted as $C_{\text{search}}(n_p)$, we select a subset of size: 
\begin{equation}
    \hat{N}_{\text{retain}}(n_p) = \min (N_{\text{retain}}, \lvert C_{\text{search}}(n_p) \rvert) \nonumber
\end{equation}
These retained children across all parents collectively constitute the next-layer parent set $M(d)$. To preserve sufficient exploration breadth $B_d$ defined in Eq. \ref{eq: branch-factor}  for each parent, we should prevent the excessive expansion of $M(d)$ through selective retention. An ideal pruning strategy is to prioritize diversity among retained nodes. Since sibling nodes share an identical prefix, highly similar intermediate reasoning steps are likely to induce similar continuations in subsequent layers, resulting in redundant chains. Such redundancy reduces effective exploration under a limited rollout budget. Motivated by this observation, we adopted a similarity-based pruning strategy over \textit{search} children.

\paragraph{Similarity-based Pruning} 
Our premise is that semantic intent of a \textit{search} action can be captured by the information it retrieves. Hence, for any two sibling \textit{search}-action nodes, $n_i$ and $n_j$, we quantify their similarity using the \textbf{Jaccard Similarity} of their top-$\text{K}$ retrieved passage sets: 
\begin{equation}
    J(i,j) = \frac{\lvert P_i  \cap P_j \rvert}{\lvert P_i  \cup P_j \rvert} \nonumber
\end{equation}
This metric ranges from $0$ to $1$. 
A larger value indicates redundant information-seeking, while a value near 0 suggests exploration of distinct knowledge pathways.
Armed with this metric, we perform pruning independently for children under each parent. Specifically, for the candidate set $C_{\text{search}}(n_p)$, we compute pairwise Jaccard distances $\delta(i,j) = 1-J(i,j)$, and apply \textbf{hierarchical clustering} to partition the nodes into $\hat{N}_{\text{retain}}(n_p)$ clusters. From each cluster, a single representative node is retained for expansion to the next layer, while the remaining nodes are pruned. This ensures that the retained nodes are diverse in terms of their information-gathering behavior, and hence likely to induce distinct future trajectories. By pruning for semantic redundancy, we ensure that our fixed computational budget is allocated to exploring broader search space, which yields more robust signals for process supervision. Crucially, this similarity-based pruning is performed online and relies solely on retrieval results for around $N$ nodes at each depth, requiring no auxiliary scoring models or annotations.

\subsection{Process Supervision}
\label{method: process-supervision}
This section details how we leverage the tree structure to transform the sparse outcome rewards into dense, step-wise advantages, thereby solving the credit assignment problem without intermediate annotations or a learned value model. 

\paragraph{Value Estimation}
We adopt the same rule-based outcome reward with exact match score as in \citet{jin2025searchr1trainingllmsreason}, which is assigned to the leaves:
\begin{equation}
    \mathcal{R}(y)=\texttt{EM}(a_{pred},a_{gold}) \nonumber
\end{equation} 
where $\mathcal{R}(y)=1$ if the extracted answer $a_{pred}$ from response $y$ matches the ground truth, and $0$ otherwise.
Rather than learning a parametric value function, we employ \textbf{Monte Carlo (MC) estimation} to back-calculate the value of any node $n_i$ from these terminal outcomes. Formally, let $L(n_i)$ be the set of all descendant leaves of $n_i$. The value $V(n_i)$ is defined as the empirical average of the outcome rewards of its descendant leaf nodes:
\begin{equation}
    V(n_i) = \frac{1}{\lvert L(n_i) \rvert} \sum_{j\in L(n_i)} \mathcal{R}_j \nonumber
\end{equation}
Intuitively, $V(n_i)$ reflects the estimated probability that continuing from step $n_i$ will eventually lead to a correct answer under current policy.

\paragraph{Advantage Calculation}
Inspired by recent work in tree-based credit assignment \cite{hou2025treerlllmreinforcementlearning,tran2025exploitingtreestructurecredit}, we define process-level supervision in terms of advantages using these estimated node values, capturing how beneficial a specific step is relative to appropriate baselines. 

Global advantage measures a step's quality relative to the overall performance on the question:
\begin{equation}
    A_{global}(n_i)=V(n_i)-V(\texttt{root}) \nonumber
\end{equation}
where $V(\texttt{root})$ is the average expected rewards over all retained trajectories for the query. 

Local advantage isolates the relative contribution of a step with respect to its immediate predecessor:
\begin{equation}
    A_{local}(n_i)=V(n_i)-V(p(n_i)) \nonumber
\end{equation}
This 
reflects whether the transition to $n_i$ improves or degrades the expected outcome.

The final process advantage of a step is the normalized sum of its global and local advantages:
\begin{equation}
    A(n_i) = \frac{1}{\sqrt{\lvert L(n_i) \rvert}} [2 \cdot V(n_i)-V(\texttt{root})-V(p(n_i))] \nonumber
\end{equation}
The scaling factor $\lvert L(n_i) \rvert^{-1/2}$ is to prevent overfitting by down-weighting the advantage of non-leaf nodes based on the size of their subtree, as these nodes may appear in multiple rollout trajectories and thus be repeatedly computed.

\begin{table*}[ht]
\centering
\scalebox{0.81}{

\begin{tabular}{lccclcll}
\toprule \toprule
\multirow{2}{*}{\textbf{Methods}} & \multicolumn{4}{c}{\textbf{Multi-hop QA}} & \multicolumn{3}{c}{\textbf{General QA}} \\ \cline{2-8} 
 & \textbf{HotpotQA} & \multicolumn{1}{c}{\textbf{2Wiki}} & \multicolumn{1}{c}{\textbf{MusiQue}} & \multicolumn{1}{c}{\textbf{Bamboogle}} & \textbf{TriviaQA} & \multicolumn{1}{c}{\textbf{PopQA}} & \multicolumn{1}{c}{\textbf{NQ}} \\
\midrule
\multicolumn{2}{l}{\textit{\textbf{Qwen2.5-7B-Instruct}}} & \multicolumn{1}{l}{} & \multicolumn{1}{l}{} & \multicolumn{1}{l}{} & \multicolumn{1}{l}{} & \multicolumn{1}{l}{} & \multicolumn{1}{l}{} \\
Rejection Sampling$^\dagger$ \cite{jin2025searchr1trainingllmsreason} & 0.331 & 0.296 & 0.123 & 0.355 & 0.592 & 0.380 & 0.360 \\
Search-R1-GRPO$^\dagger$ \cite{jin2025searchr1trainingllmsreason} & 0.386 & 0.346 & 0.162 & 0.400 & 0.623 & 0.427 & 0.429 \\
Search-R1-PPO$^\dagger$ \cite{jin2025searchr1trainingllmsreason} & 0.370 & 0.414 & 0.146 & 0.368 & 0.610 & 0.397 & 0.393 \\
ReasonRAG$^\dagger$ \cite{zhang2025process} & 0.384 & 0.436 & 0.128 & 0.360 & - & 0.415 & - \\
StepSearch$^\dagger$ \cite{zheng-etal-2025-stepsearch}& 0.386 & 0.366 & 0.226 & 0.400 & - & - & - \\
GiGPO$^\dagger$ \cite{feng2025groupingroup}&  0.416 & 0.436 & 0.189 & 0.408$^*$ & 0.647 & 0.461 & 0.464 \\
\rowcolor{gray!15}
\textbf{Ours}  &  \textbf{0.507} & \textbf{0.482} & \textbf{0.239} & \textbf{0.520} & \textbf{0.676} & \textbf{0.485} & \textbf{0.521} \\
\midrule 

\multicolumn{2}{l}{\textit{\textbf{Qwen2.5-3B-Instruct}}} & \multicolumn{1}{l}{} & \multicolumn{1}{l}{} & \multicolumn{1}{l}{} & \multicolumn{1}{l}{} & \multicolumn{1}{l}{} & \multicolumn{1}{l}{} \\
Rejection Sampling$^\dagger$ \cite{jin2025searchr1trainingllmsreason} & 0.240 & 0.233 & 0.059 & 0.210 & 0.488 & 0.332 & 0.294 \\
Search-R1-GRPO$^\dagger$ \cite{jin2025searchr1trainingllmsreason} & 0.331 & 0.310 & 0.124 & 0.232 & 0.565 & 0.391 & 0.397 \\
Search-R1-PPO$^\dagger$ \cite{jin2025searchr1trainingllmsreason} & 0.324 & 0.319 & 0.103 & 0.264 & 0.545 & 0.378 & 0.341 \\
ReasonRAG$^\dagger$ \cite{zhang2025process} & 0.300 & 0.266 & 0.069 & 0.136 & - & 0.329 & - \\
StepSearch$^\dagger$ \cite{zheng-etal-2025-stepsearch} & 0.345 & 0.320 & 0.174 & 0.344 & - & - & - \\
GiGPO$^\dagger$  \cite{feng2025groupingroup} & 0.369 & 0.370 & 0.126 & 0.304$^*$ & 0.595 & 0.424 & 0.420 \\
\rowcolor{gray!15}
\textbf{Ours}  &  \textbf{0.462} & \textbf{0.456} & \textbf{0.215} & \textbf{0.440} & \textbf{0.636} & \textbf{0.460} & \textbf{0.476} \\
\midrule \midrule

\multicolumn{2}{l}{\textit{\textbf{Qwen3-8B (no thinking)}}} & \multicolumn{1}{l}{} & \multicolumn{1}{l}{} & \multicolumn{1}{l}{} & \multicolumn{1}{l}{} & \multicolumn{1}{l}{} & \multicolumn{1}{l}{} \\
Search-R1-GRPO & 0.468 & 0.503 & 0.218 & \textbf{0.520} & 0.670 & 0.462 & 0.452 \\
Search-R1-PPO & 0.452 & 0.499 & 0.213 & 0.504 & 0.674 & 0.450 & 0.461 \\
ReasonRAG$^*$ & 0.388 & 0.450 & 0.142 & 0.392 & - & 0.274 & - \\
GiGPO & 0.411 & 0.450 & 0.175 & 0.432 & 0.662 & 0.431 & 0.458 \\
\rowcolor{gray!15}
\textbf{Ours} & \textbf{0.487} & \textbf{0.514} & \textbf{0.238} & \textbf{0.520} & \textbf{0.687} & \textbf{0.477} & \textbf{0.489} \\
\midrule

\multicolumn{2}{l}{\textit{\textbf{Qwen3-4B-Instruct-2507}}} & \multicolumn{1}{l}{} & \multicolumn{1}{l}{} & \multicolumn{1}{l}{} & \multicolumn{1}{l}{} & \multicolumn{1}{l}{} & \multicolumn{1}{l}{} \\
Search-R1-GRPO & 0.474 & 0.517 & 0.225 & \textbf{0.536} & 0.675 & 0.462 & 0.447 \\
Search-R1-PPO & 0.454 & 0.507 & 0.215 & 0.480 & 0.662 & 0.451 & 0.433 \\
GiGPO & 0.424 & 0.428 & 0.184 & 0.472 & 0.643 & 0.467 & 0.449 \\
\rowcolor{gray!15}
\textbf{Ours} & \textbf{0.480} & \textbf{0.541} & \textbf{0.233} & \textbf{0.536} & \textbf{0.680} & \textbf{0.488} & \textbf{0.476}
\\ \bottomrule \bottomrule
\end{tabular}
}

\caption{\textbf{Main Results.} $^\dagger$ indicates results cited from the original work. The best performance is set in \textbf{bold}. $^*$ denotes specially reproduced results with details in Appendix \ref{appendix: baseline}. For online rollout RL methods, i.e., Ours, Search-R1-GRPO/PPO, and GiGPO, we train all models under a unified setting on Qwen3 series for fair comparison: using $12\text{K}$ randomly sampled training instances from HotpotQA and NQ with rollout group size $N=8$.
For Qwen2.5 series, the original works of Search-R1 and GiGPO trained models with $N=5$ on full HotpotQA and NQ training split ($170\text{K}$). Methods relying on intermediate labels or offline process-supervised data are mainly reported using source results, as their training data and supervision protocols are not directly comparable: ReasonRAG trained models with $5\text{K}$ questions sampled PopQA, HotpotQA, and 2WikiMultihopQA, while StepSearch augments the MusiQue dataset with \texttt{GPT-4o} generated intermediate labels.}

\label{tab: main-results}
\end{table*}
\begin{table*}[ht]
\centering
\scalebox{0.75}{
\begin{tabular}{lccccccc}
\toprule
\textbf{Methods} & \textbf{HotpotQA} & \textbf{2Wiki} & \textbf{MusiQue} & \textbf{Bamboogle} & \textbf{TriviaQA} & \textbf{PopQA} & \textbf{NQ} \\
\midrule
Search-R1-GRPO & 0.474 & 0.517 & 0.225 & 0.536 & 0.675 & 0.462 & 0.447 \\
Ours & \textbf{0.480} & \textbf{0.541} & \textbf{0.233} & \textbf{0.536} & \textbf{0.680} & \textbf{0.488} & \textbf{0.476} \\
  \quad w/o process advantage & 0.463 & 0.523 & 0.216 & 0.472 & 0.674 & 0.470 & 0.454 \\
  \quad w/o process advantage \& similarity-based pruning & 0.430 & 0.509 & 0.211 & 0.464 & 0.677 & 0.463 & 0.450 \\
\bottomrule
\end{tabular}
}
\caption{Ablation study on the effectiveness of tree-based process advantage with Qwen3-4B-Instruct-2507.}
\label{tab: ablation-study}
\end{table*}

\paragraph{Policy Optimization} To derive online training data, we first convert the pruned tree into a set of trajectories. Specifically, each leaf node corresponds to the end of a complete rollout. Backtracking from the leaf to the root yields 
a full agentic trajectory, as a sequence of steps. From the collection of all such paths, we randomly sample $N$ trajectories as experience to form the training set $Y=\{y_1, \dots, y_N\}$ for policy optimization. Under this formulation, a single tree node may appear as a step in multiple trajectories, and its associated process advantage is shared across all such occurrences.

For optimization, the node advantage $A(n_i)$ is broadcast to all language model-generated tokens $(r_i,a_i)$ within that step.
Following prior work \cite{jin2025searchr1trainingllmsreason,song2025r1searcherincentivizingsearchcapability}, we apply a loss mask to the observation tokens (retrieved passages), ensuring that gradients are computed only over the agent's output. This produces a token-level training signal that reflects the step-wise credit assignment induced by the tree. The resulting objective in Appendix \ref{appendix:objective} remains fully compatible with standard policy gradient methods, differing only in that our process-level advantages are integrated.

\section{Experiments}

\paragraph{Datasets \& Metrics} Following \citet{jin2025searchr1trainingllmsreason}, we evaluate \textsc{TreePS-RAG} on seven QA benchmarks, categorized as follows: (1) three single-hop QA datasets, including NQ \cite{kwiatkowski-etal-2019-natural}, TriviaQA \cite{joshi-etal-2017-triviaqa}, and PopQA \cite{mallen-etal-2023-trust}, and (2) four multi-hop QA datasets, including HotpotQA \cite{yang-etal-2018-hotpotqa}, 2WikiMultihopQA \cite{ho-etal-2020-constructing}, Bamboogle \cite{press-etal-2023-measuring}, and MuSiQue \cite{trivedi-etal-2022-MuSiQue}. We use Exact Match (EM) as the primary evaluation metric across all benchmarks.
\paragraph{Baselines} We evaluate \textsc{TreePS-RAG} against a set of competitive and representative training baselines.
We include (1) Search-R1 \cite{jin2025searchr1trainingllmsreason}, our primary baseline, that trains models using GRPO and PPO under outcome-only supervision. We further compare against several recent methods that incorporate step-level supervision: (2) StepSearch \cite{zheng-etal-2025-stepsearch} leverages step-wise PPO by training on MuSiQue dataset augmented with intermediate subquestion-level annotations, which are generated by \texttt{GPT-4o} to provide process-level rewards. (3) ReasonRAG \cite{zhang2025process} constructs an offline process-supervised dataset via Monte Carlo Tree Search (MCTS) exploration and subsequently applies DPO to optimize the policy on this fixed dataset. (4) GiGPO \cite{feng2025groupingroup} augments episode-level GRPO by introducing an anchor-based grouping mechanism over repeated environment states, enabling step-level credit assignment during training. See implementation details in Appendix \ref{appendix: implementation}.

\section{Main Results}

As shown in Table \ref{tab: main-results}, \textbf{\textsc{TreePS-RAG} consistently outperforms all competitive training baselines across seven QA benchmarks}, achieving average performance of $49.0\%$, $44.9\%$, $48.7\%$ and $49.0\%$ with Qwen2.5-3B-Instruct, Qwen2.5-7B-Instruct, Qwen3-8B (no-thinking mode) and Qwen3-4B-Instruct-2507. The performance gains remain stable across four backbone models under both in-distribution and out-of-domain evaluation, indicating robustness and generalization of our approach. 

Compared to outcome-supervised baselines Search-R1-GRPO/PPO, \textsc{TreePS-RAG} achieves consistent improvements. On the Qwen3 series, all online RL methods are trained under identical rollout budgets, data splits, and optimization settings, differing only in the supervision. Consequently, the observed gains can be attributed to our tree-derived, step-wise process advantages, which provide informative and fine-grained credit assignment during policy optimization. In contrast, the baselines assign supervision solely based on the final outcome, lacking the ability to distinguish beneficial intermediate decisions from suboptimal ones. 

Beyond outperforming outcome-only baselines, \textsc{TreePS-RAG} also surpasses process-supervised approaches. 
ReasonRAG relies on offline-constructed training data,
while StepSearch injects intermediate supervision through \texttt{GPT-4o} pre-generated annotations. In comparison with them, our method achieves stronger performance while remaining fully online and free of intermediate labeling. This highlights the contribution of our approach: (1) effective process supervision can be derived directly from outcome rewards through tree structure, avoiding costly step-level annotation; (2) online RL paradigm to avoid distribution shift of offline training \cite{shenfeld2025rlsrazoronlinereinforcement,chen2025retainingdoingroleonpolicy}. 

Finally, under identical configuration, \textsc{TreePS-RAG} consistently exceeds GiGPO on the Qwen3 series, which also aims to improve step-level credit assignment through anchor-based grouping. This comparison isolates the benefits of our tree-structured supervision: explicitly modeling the rollout space as a tree and estimating step utility via Monte Carlo returns yields a more informative and reliable process signal. Overall, these results demonstrate that \textbf{online tree-based process supervision enables effective credit assignment for agentic RAG training}, improving performance without introducing additional annotations.

\section{Analysis}
\begin{table*}[ht]
\centering
\scalebox{0.8}{
\begin{tabular}{lcccccccc}
\toprule
\textbf{Tree} & \textbf{HotpotQA} & \textbf{2Wiki} & \textbf{MuSiQue} & \textbf{Bamboogle} & \textbf{TriviaQA} & \textbf{PopQA} & \textbf{NQ} & \textbf{Avg} \\
\midrule
Default (Dynamic Tree) & 0.480 & 0.541 & 0.233 & 0.536 & 0.680 & 0.488 & 0.476 & 0.490 \\
Larger Tree & 0.490 & 0.531 & 0.239 & 0.536 & 0.688 & 0.487 & 0.494 & 0.495 \\
\bottomrule
\end{tabular}}
\caption{Effect of tree scale. Default corresponds to our dynamic branching strategy defined in Eq. \ref{eq: branch-factor} with $N_{\text{retain}}=2$ and $D=4$, which results in approximate $[8,4,2,1]$ from depth $1$ to $4$. Larger Tree increases the branching factor to fixed $[9,7,5,1]$ and raises $N_{\text{retain}}=3$. $N=8$ trajectories are sampled during RL training for both variants.}
\label{table: tree-scale}
\end{table*}

\subsection{Effect of Tree-based Process Supervision}

To disentangle the contributions of our core components, we conduct a detailed ablation study with results summarized in Table \ref{tab: ablation-study}. We compare four variants, all trained with the same group size $N=8$: 
(1) the competitive primary baseline Search-R1-GRPO, 
(2) our full method (Ours), 
(3) Ours w/o process advantage (PA), which retains the proposed tree-based trajectory construction but reverts to standard GRPO objective and advantage definition used by Search-R1, and (4) Ours w/o process advantage (PA) \& similarity-based pruning (SP), which further removes our similarity-based pruning and instead randomly retains $ \hat{N}_{\text{retain}}(n_p)$ nodes during tree expansion.

\paragraph{Efficacy of Tree-based Trajectory Construction} 
Surprisingly,  Ours w/o PA achieves performance comparable to Search-R1-GRPO across most datasets, despite sampling trajectories from a shared-prefix tree structure. In principle, such shared-prefix rollouts could reduce trajectory diversity compared to independently sampling in parallel,
potentially degrading learning under standard GRPO. The absence of noticeable performance degradation suggests that \textbf{our online tree construction, together with selective pruning, effectively preserves sufficient exploration diversity under a comparable computational budget}.

\paragraph{Efficacy of Similarity-based Pruning} Removing selective pruning leads to a clear and consistent performance drop. Ours w/o PA \& SP underperforms both Ours w/o PA and Search-R1-GRPO on most datasets, with especially pronounced degradation on multi-hop benchmarks. This indicates that naively selecting nodes without carefully controlling redundancy is insufficient. 

\paragraph{Impact of tree-based process advantage} Most importantly, comparing the full mode (Ours) with  Ours w/o PA  reveals a consistent and notable improvement across all datasets. This performance lift can be directly and unambiguously attributed to the effectiveness of our tree-derived process advantage. By providing a fine-grained, step-wise credit assignment signal, \textbf{our method enables more precise and reliable policy optimization} than the uniform, trajectory-level feedback of standard GRPO.

\subsection{Effect of Tree Scale}
We analyze the effect of tree scale by increasing both the branching factor $B_d$ and the retained child nodes per parent $N_{\text{retain}}$. As shown in Table \ref{table: tree-scale}, enlarging the tree leads to a consistent but modest improvement in average performance ($0.49 \rightarrow 0.495$). This is expected under our controlled experimental setup. During RL training, all variants sample the same number trajectories ($N=8$) as experience from the constructed tree for policy optimization to ensure fair comparison. Consequently, increasing the tree size does not increase the amount of policy optimization signal, but primarily improves the quality of Monte Carlo estimation for the process advantage by aggregating over more leaf nodes.
These results suggest that our default dynamic branching strategy already strikes a favorable balance between exploration diversity and computational efficiency. Further enlarging the tree mainly reduces estimation variance, leading to consistent but bounded performance gains.

\subsection{Continuation-Based Robustness Analysis}
\begin{figure}[ht]
\centering
\includegraphics[width=0.48\textwidth]{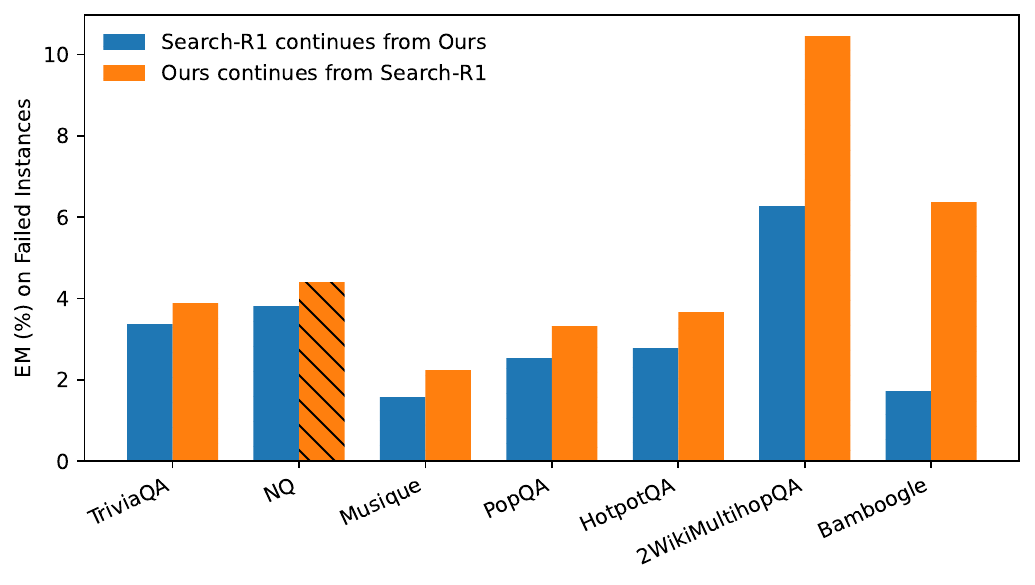}
\caption{Analysis of continuation-based reasoning.}
\label{fig: analysis-continuation}
\end{figure}
\noindent While final accuracy reflects end-to-end performance, it does not directly reveal the quality of reasoning steps. In the absence of step-level annotations, we introduce a continuation-based analysis to assess whether our process-supervised training leads to more reliable reasoning steps. We compare \textsc{TreePS-RAG} with Search-R1-GRPO by probing a model’s ability to recover from imperfect reasoning prefixes. 
For each dataset, we collect all instances where one method fails to produce correct final answers. We truncate these failed trajectories by removing the final \textit{answer} step, and use the remaining prefix as the input context. The other method is prompted to continue reasoning from this identical prefix. This isolates the model’s ability to interpret, verify, and correct an existing reasoning trajectory, rather than solving the problem from scratch.
As shown in Figure \ref{fig: analysis-continuation}, across all datasets, \textbf{ \textsc{TreePS-RAG} consistently yields larger performance gains when continuing from Search-R1-generated prefixes than the reverse setting.} This asymmetry suggests that our method is more effective at reflecting intermediate steps and correcting earlier mistakes, even when starting from imperfect trajectories. This finding complements our main results by providing direct evidence that \textsc{TreePS-RAG} yields higher-quality reasoning processes beyond improved final accuracy.

\paragraph{Case Study} 
Figure \ref{fig: case-study-1} and \ref{fig: case-study-2} provide qualitative examples that illustrate \textbf{how process supervision improves step-wise reasoning} under continuation, including both search and answer actions. Across both cases, Search-R1-GRPO fails to recover from imperfect prefixes, either by grounding answers to misleading evidence or by prematurely terminating reasoning under incomplete information. In contrast, \textsc{TreePS-RAG} identifies missing or unreliable evidence, and revises the trajectory through additional verification with rewritten query. 
See detailed analysis in Appendix \ref{appendix: case-study}.



\section{Related Works}
\paragraph{Agentic RAG.}
Early agentic RAG methods mainly relied on prompt engineering and self-refinement mechanisms \cite{madaan2023selfrefineiterativerefinementselffeedback,gou2024criticlargelanguagemodels}, which includes interleaving retrieval with LLM reasoning-based generation iteratively to solve complex knowledge-intensive questions \cite{trivedi2023interleavingretrievalchainofthoughtreasoning,jiang-etal-2023-active,shao2023enhancingretrievalaugmentedlargelanguage}, and adaptive rag that encourages LLMs to decide when and how to retrieve \cite{jeong-etal-2024-adaptive,wang-etal-2025-self}. 
Recent methods advanced towards trainable agents such as Self-RAG \cite{asai2023selfraglearningretrievegenerate} and Auto-RAG \cite{yu2024autoragautonomousretrievalaugmentedgeneration}. 
Methods \cite{jin2025searchr1trainingllmsreason, song2025r1searcherincentivizingsearchcapability, chen2025researchlearningreasonsearch, sun2025zerosearchincentivizesearchcapability} applied GRPO \cite{shao2024deepseekmathpushinglimitsmathematical} or PPO \cite{schulman2017proximalpolicyoptimizationalgorithms}-based reinforcement learning outcome-only supervision to train search-capable agents. While effective, these methods relying on sparse, delayed rewards limit fine-grained credit assignment. This motivates the exploration of process-supervised RL such as ReasonRAG \cite{zhang2025process}, GiGPO \cite{feng2025groupingroup} and StepSearch \cite{zheng-etal-2025-stepsearch}, to provide step-wise guidance during policy learning.

\paragraph{Tree Structure.}
Tree-structured reasoning \cite{huang2025treeopooffpolicymontecarlo} has been extensively studied in LLM research, including data synthesis, offline alignment, and enhanced credit assignment. Previous works 
\cite{zhang2025process,xie2024montecarlotreesearch}  employ Monte Carlo Tree Search (MCTS) to generate step-level supervision signals or preference pairs for downstream training, such as DPO \cite{rafailov2024directpreferenceoptimizationlanguage}.
More recent studies exploit the tree structure of rollouts to densify credit assignment during RL training on math problems, such as TreePO \cite{li2025treepobridginggappolicy}, TreeRPO \cite{yang2025treerpotreerelativepolicy}, and TreeRL \cite{hou2025treerlllmreinforcementlearning}.

\section{Conclusion}
We present \textbf{\textsc{TreePS-RAG}}, an online tree-based RL framework for agentic RAG that enables step-wise credit assignment without intermediate annotations or auxiliary reward or value models. \textsc{TreePS-RAG} builds a rollout tree and uses MC estimates from outcome-supervised rollouts to compute process advantages for intermediate steps, while retaining the simplicity of standard outcome-based optimization. With efficient depth-wise branching and similarity-based pruning, it preserves exploration diversity at comparable cost to outcome-only training. Experiments on seven QA benchmarks across multiple model scales show consistent improvements over strong RL baselines.

\section*{Limitations}
Despite the promising results, our work has several limitations that need to be addressed in future work.
First, due to limited computational resources, we primarily conduct experiments on small- to medium-scale backbone models. Training RL-based methods requires repeated policy rollouts and optimization steps, which makes large-scale training resource-intensive. While our approach is model-agnostic, scaling it to much larger language models would require additional computational resources and further engineering optimization. 
Second, our evaluation is limited to English, text-only QA benchmarks. Extending our approach to multilingual or multimodal settings remains an important direction for future work. Finally, while tree-based rollouts introduce additional coordination complexity compared to fully independent sampling, we mitigate this through an efficient online tree construction strategy that operates under a comparable rollout budget with standard outcome-supervised approaches. In addition, we implement asynchronous rollout and sampling during RL training, which significantly improves throughput and keeps the overall training cost within a manageable range. Further reducing latency and improving system efficiency is left for future work.

\section*{Ethical Considerations}
While our proposed RL framework for agentic RAG demonstrates strong performance, the deployment of such a system requires careful attention to content safety. Since the system's outputs are conditioned on external data, the potential presence of offensive or biased material can lead to unintended harmful generation. We strongly advise users to mitigate potential risks by ensuring that all performance assessments utilize curated, non-toxic benchmarks.

\bibliography{custom}

\appendix

\label{sec:appendix}

\section{Prompt Template}
\begin{figure*}[ht]
\centering
\includegraphics[width=1\textwidth]{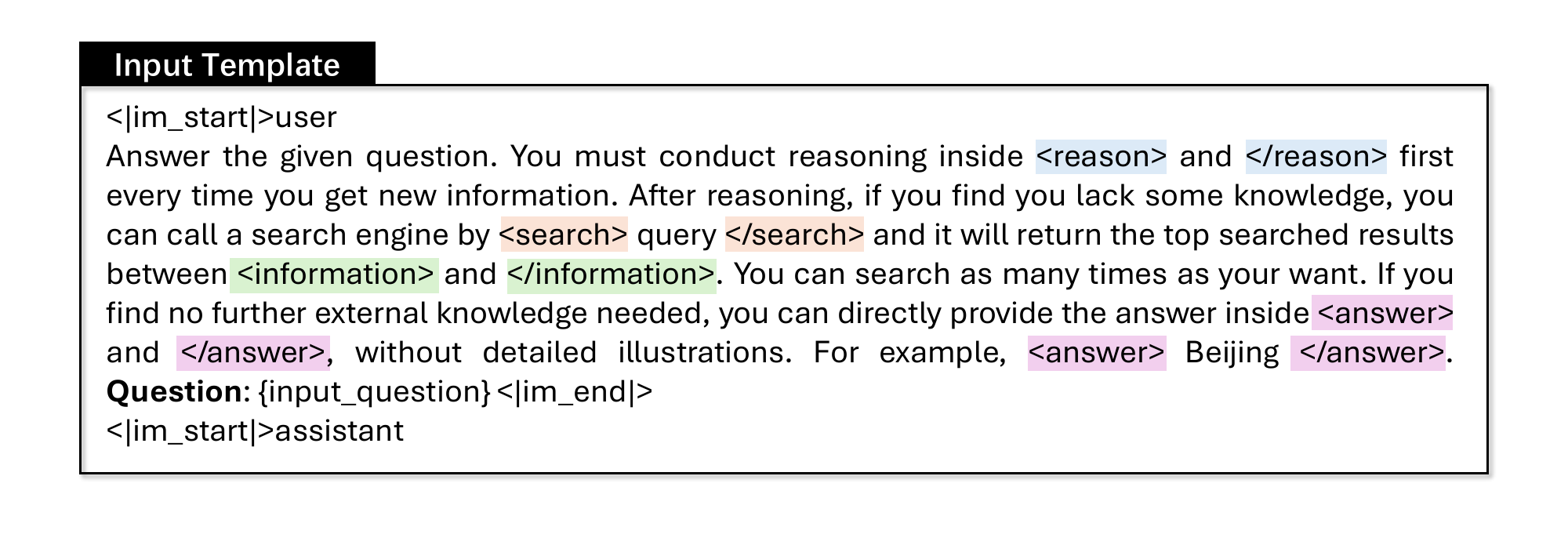}
\caption{Template for \textsc{TreePS-RAG}. \{\texttt{input\_question}\} will be replaced with specific question during training and inference. We follow \citet{jin2025searchr1trainingllmsreason} and only replace \texttt{<think>} with \texttt{<reason>} to adapt for Qwen3 models.}
\label{fig: input-prompt}
\end{figure*}

Figure~\ref{fig: input-prompt} shows the input prompt template used during both training and inference. Our template largely follows that of Search-R1 \cite{jin2025searchr1trainingllmsreason}, with a minimal modification to accommodate Qwen3 models. Specifically, we replace the \texttt{<think>} token used in Search-R1 with \texttt{<reason>}. This change is motivated by differences in tokenizer across Qwen model families. In the Qwen3 series, \texttt{<think>} is treated as a special token associated with the model’s thinking mode. Qwen3-4B-Instruct-2507 is not explicitly trained to use \texttt{<think>} as a control signal, while Qwen3-8B uses it to support its pretrained native thinking interface. To ensure consistent behavior across model variants and to avoid entangling our agentic RAG reasoning traces with model-specific thinking modes, we adopt \texttt{<reason>} as a replacement.

\section{Training Objective}
\label{appendix:objective}
Our training objective remains fully compatible with standard policy gradient methods, differing only in that our dense, process-level advantages defined in \S \ref{method: process-supervision} are integrated.

\begin{equation}
\begin{multlined}[0.9\columnwidth]
\mathcal{J}(\theta)
= \mathbb{E}_{x,\{y_i\}_{i=1}^{N} \sim \pi_{\theta_{\text{old}}}(\cdot \mid x)}
\Bigg[
\frac{1}{\sum_{i=1}^{N} |y_i|}
\sum_{i=1}^{N}
\sum_{t=1}^{|y_i|} \\
 \min\!\Big(
\frac{
\pi_\theta\!\left(y_{i,t} \mid x, y_{i,<t}\right)
}{
\pi_{\theta_{\mathrm{old}}}\!\left(y_{i,t} \mid x, y_{i,<t}\right)
}\,A_{i,t}, \\
\operatorname{clip}\!\big( 
\frac{
\pi_\theta\!\left(y_{i,t} \mid x, y_{i,<t}\right)
}{
\pi_{\theta_{\mathrm{old}}}\!\left(y_{i,t} \mid x, y_{i,<t}\right)
},
1-\epsilon,
1+\epsilon
\big)\,A_{i,t}
\Big) \\
- \beta \, D_{\mathrm{KL}}\!\left[
\pi_\theta \middle\|\;
\pi_{\text{ref}}
\right]
\Bigg]
\end{multlined}
\label{eq: objective}
\end{equation}

\begin{algorithm*}[ht]
\caption{\textsc{TreePS-RAG}: Online Tree Construction and Process Advantage Calculation}
\label{alg:main}
\begin{algorithmic}[1]
\Require Question $q$, ground-truth answer $a_{gold}$, rollout budget $N$, depth limit $D$, local retention budget $N_{retain}$, large language model $\pi_\theta$, retriever $\mathcal{R}$, passage number $K$.
\Ensure A pruned tree $\mathcal{T}$ where each node $n_i$ is annotated with a process advantage $A(n_i)$.

\Statex \Comment{\colorbox[HTML]{ECF4FF}{\textit{Phase 1: Online Tree Construction}}}
\State Initialize tree $\mathcal{T}$ with root node $n_{\texttt{root}}(q)$
\State $M(0) \gets \{n_{\texttt{root}}\}$ \Comment{Initialize the set of nodes to be expanded at depth 0.}

\For{$d = 1$ to $D$} 
    \State $M(d) \gets \emptyset$ \Comment{Initialize the set of retained nodes for the current depth}
    \State $B_d \gets \lceil N / |M(d-1)| \rceil$  \Comment{Update branching factor for expansion}
    \For{each parent node $n_p \in M(d-1)$}
        \State $C(n_p) \gets \{n_p^1, \dots, n_p^{B_{d}}\} \sim \pi_\theta$ \Comment{Sample $B_{d}$ children from parent $n_p$}
        
        \State $C_{\text{search}}(n_p) \gets \{n_p^i \in C(n_p) \mid a_{n_p^i}=\text{search}\}$ \Comment{Identify searchable children}
        \State $C_{\text{answer}}(n_p) \gets \{n_p \in C(n_p) \mid a_{n_p^i}=\text{answer}\}$ \Comment{Identify terminal answer children}
        
        \State $\hat{N}_{\text{retain}}(n_p) \gets \min (N_{\text{retain}}, \lvert C_{\text{search}}(n_p) \rvert)$
        \For{each search child $n_p^i \in C_{\text{search}}(n_p)$}
            \State $P_i \gets \mathcal{R}(q_i, K)$ \Comment{Retrieve passages for similarity calculation}
        \EndFor
        \State $C_{\text{retain}}(n_p) \gets \mathcal{F}_{prune}(C_{\text{search}}(n_p), \hat{N}_{\text{retain}}(n_p) )$ \Comment{Similarity-based clustering}

        \State $M(d) \gets M(d) \cup C_{\text{retain}}(p)$ \Comment{Add locally retained nodes to the global set}
        \State $\mathcal{T} \gets \mathcal{T} \cup C_{\text{retain}} \cup C_{\text{answer}}(p) $ \Comment{Add retained search and answer children to the tree}
    \EndFor
\EndFor 

\Statex \Comment{\colorbox[HTML]{ECF4FF}{\textit{Phase 2: Process Advantage Calculation}}}
\For{each node $n_i \in \mathcal{T}$} 
    \If{$a_{n_i} = \texttt{answer}$ OR depth$(n_i) = D$}
        \State $r \gets \texttt{EM}(a_{\text{pred}}, a_{\text{gold}})$ \Comment{Assign outcome rewards to leaves}
    \EndIf
\EndFor

\For{each node $n_i \in \mathcal{T}$}
    \State $V(n_i) \gets \frac{1}{|L(n_i)|} \sum_{n_j \in L(n_i)} r(n_j)$   \Comment{Calculate node values}
    \State $A(n_i) \gets \frac{1}{\sqrt{|L(n_i)|}} \left[ 2 \cdot V(n_i) - V(n_{\texttt{root}}) - V(p(n_i)) \right]$ \Comment{Calculate process advantages}
\EndFor

\end{algorithmic}
\end{algorithm*}
\section{Implementation Details}
\label{appendix: implementation}

\subsection{Datasets}
Table \ref{tab: data-statistics} shows the data statistics. The $12\text{K}$ training set is used for all our implemented online RL approaches, which include Ours, Search-R1 (GRPO \& PPO), and GiGPO.
\begin{table}[ht]
\centering
\scalebox{0.9}{
\begin{tabular}{lcc}
\toprule
\textbf{Dataset} & \textbf{Train Set Size} & \textbf{Test Set Size} \\
\midrule
HotpotQA & 10,000 & 7,405 \\
2WikimultihopQA & - & 12,576 \\
MuSiQue & - & 2,417 \\
Bamboogle & - & 125 \\
TriviaQA & - & 11,313 \\
PopQA & - & 14,267 \\
NQ & 2,000 & 3,610 \\
\midrule
\textbf{Total} & \textbf{12,000} & \textbf{51,713} \\
\bottomrule 
\end{tabular}
}
\caption{Number of examples in our training and evaluation sets.}
\label{tab: data-statistics}
\end{table}

\subsection{Training Details}
The experiments in this paper are conducted using VeRL \cite{10.1145/3689031.3696075}, a open-source library for LLM reinforcement learning. The rollout operates in SGLang \cite{zheng2024sglangefficientexecutionstructured} inference engine under asynchronous settings to speed up the training. The training runs on 8 H20 GPUs and takes approximately 68 hours. The hyperparameters for the training are listed in Tab. \ref{tab:hyperpara}. During inference, we set the maximum number of action steps as $64$ for  our approach  and all reproduced approaches (Search-R1-GRPO/PPO and GiGPO).

\begin{table}[ht]
\centering
\scalebox{0.75}{
\begin{tabular}{ll}
\toprule
\textbf{Hyperparameters}  \\ \hline
Training batch size & 512 \\
Optimizer & AdamW \\
 &  \cite{Loshchilov2017DecoupledWD}\\
Learning rate & 1e-6 \\
Warmup ratio & 0.285 \\
Gradient accumulation step & 1 \\
Learning rate scheduler & Linear \\
KL coefficient & 0.001 \\
Temperature & 1.0 \\ 
Maximum length (per turn) & 512 \\ 
Total epoch & 10 \\ 
Rollout group size & 8 \\ 
Maximum step & 4 \\
\bottomrule 
\end{tabular}}
\caption{The hyperparameters used in our RL training.}
\label{tab:hyperpara}
\end{table}

\subsection{Retrieval}
\label{appendix: retrieval}
We follow the same setting in Search-R1 \citet{jin2025searchr1trainingllmsreason}, with \textbf{2018 Wikipedia} dump \cite{karpukhin-etal-2020-dense} as the corpus as the knowledge source and E5 \cite{wang2024textembeddingsweaklysupervisedcontrastive}  as the retriever. 
The retriever is implemented with Faiss \cite{johnson2019billion} to accelerate the retrieval process. The Top-3 passages are returned, constrained to a maximum combined length of 512 tokens.

\subsection{Baselines}
\label{appendix: baseline}
We assess \textsc{TreePS-RAG} across multiple backbone language models, including Qwen2.5-3B/7B-Instruct, Qwen3-4B-Instruct-2507, and Qwen3-8B (non-thinking mode). 
For methods that rely on \textbf{online} rollout-based reinforcement learning, including Ours, Search-R1 and GiGPO, we train all models under a unified experimental setting to ensure fair comparison with Qwen3 series. For this setting, we randomly sample $12\text{K}$ training instances from the NQ and HotpotQA training sets, and fix the rollout group size to $N=8$. NQ and HotpotQA are then used for in-domain evaluation, with the remaining five serving as an out-of-domain test. We cite the results of Search-R1 and GiGPO with Qwen2.5 series from the original source works, with rollout group size $N=5$ and on the full combination of HotpotQA and NQ training split, i.e., $170\text{K}$.
StepSearch and ReasonRAG rely on specifically-constructed process-supervised datasets, which constitute a core part of their methodological contributions.
ReasonRAG generates training \textbf{offline} trajectories over $5\text{K}$ questions sampled from PopQA, HotpotQA, and 2WikiMultihopQA. StepSearch augments the MuSiQue dataset with intermediate annotations pre-generated by \texttt{GPT-4o} as labels for process rewards. Due to the differences, 
it is non-trivial to reproduce them under a fully unified setting.
We therefore mainly report their original results. 
Below we detail two specially reproduced baselines results indicated with $^*$ symbol in Table \ref{tab: main-results}.

When reproducing GiGPO using the Qwen2.5 series, we observed an inconsistency in the reported Bamboogle evaluation. Specifically, the reported score does not appear to be normalized by the benchmark size (125), which may be related to the batch-based evaluation procedure. To ensure a fair and consistent comparison, we therefore report our own reproduced results on Bamboogle. For the remaining benchmarks, our reproduced results closely match those reported in the original paper, and we thus directly cite their original scores.

ReasonRAG is trained with a different data source, augmenting the 2018 Wikipedia dump with passages from PopQA, HotpotQA, and 2WikiMultihopQA, rather than our main experiments that using 2018 Wikipedia dump only (Appendix \ref{appendix: retrieval}). Since the offline preference training trajectories are generated using BGE\footnote{\url{https://huggingface.co/BAAI/bge-base-en-v1.5}} as the retriever with top-$3$ retrieval, we follow the same retriever during reproduction to avoid discrepancies between training and inference. Using their MCTS-constructed dataset, we are able to reproduce performance comparable to the reported results on Qwen2.5-7B-Instruct. However, when applying the same pipeline to the Qwen3 series, the reproduced performance is less competitive, particularly on PopQA. We therefore report the best results for each dataset obtained after DPO training, based on multiple hyperparameter trials, under both the BGE-indexed augmented corpus and the BGE-indexed 2018 Wikipedia dump. Further analysis suggests that this difference may stem from behavioral differences between the backbone models. Under the iteration-0 prompt designed in ReasonRAG (Table \ref{tab:reasonrag-iter0-prompt}), Qwen2.5-7B-Instruct tends to actively generate search queries and rely on retrieved passages, with only around $3\%$ of PopQA instances attempting to answer directly using parametric knowledge. In contrast, Qwen3-8B (no thinking mode) answers nearly $80\%$ of questions directly in the first iteration without invoking retrieval. While DPO training partially mitigates this tendency, it does not fully shift the model’s behavior given the limited scale of the training data ($5$K questions produced $13$K process steps).

\begin{table}[t]
\centering
\small
\setlength{\tabcolsep}{6pt}
\renewcommand{\arraystretch}{1.2}
\begin{tabular}{p{0.95\linewidth}}
\toprule
\textbf{Iteration-0 Prompt Used in ReasonRAG} \\
\midrule
\textbf{System:} You are an assistant for question answering with access to a retrieval tool. Upon receiving a question, your task is to: \\
\quad $\bullet$ Analyze and Decompose the Question: Break the question into smaller, manageable sub-questions to ensure all aspects are addressed. \\
\quad $\bullet$ Evaluate Your Knowledge: Assess each sub-question or component: \\
\quad\quad -- Identify parts you can confidently answer based on your existing knowledge. \\
\quad\quad -- Pinpoint parts that require additional information or verification through retrieval tools. \\
\quad $\bullet$ Conciseness: Ensure both queries and answers are concise, using nouns or short phrases whenever possible. \\
\quad $\bullet$ Respond Format: \\
\quad\quad If your knowledge is sufficient to answer the question, conclude with: \\
\quad\quad \texttt{``So the answer is <answer>answer</answer>''} \\
\quad\quad If retrieval is necessary to provide a complete answer, conclude with: \\
\quad\quad \texttt{``So the next query is <query>query</query>''} \\[0.5em]
\textbf{User:} \{input\_question\} \\
\bottomrule
\end{tabular}
\caption{Iteration-0 prompt used in ReasonRAG, which allows the model to answer using its parametric knowledge if deemed sufficient and confident, or to invoke retrieval otherwise.}
\label{tab:reasonrag-iter0-prompt}
\end{table}


\section{Case Study}
\label{appendix: case-study}
\begin{figure*}[ht]
\centering
\includegraphics[width=0.95\textwidth]{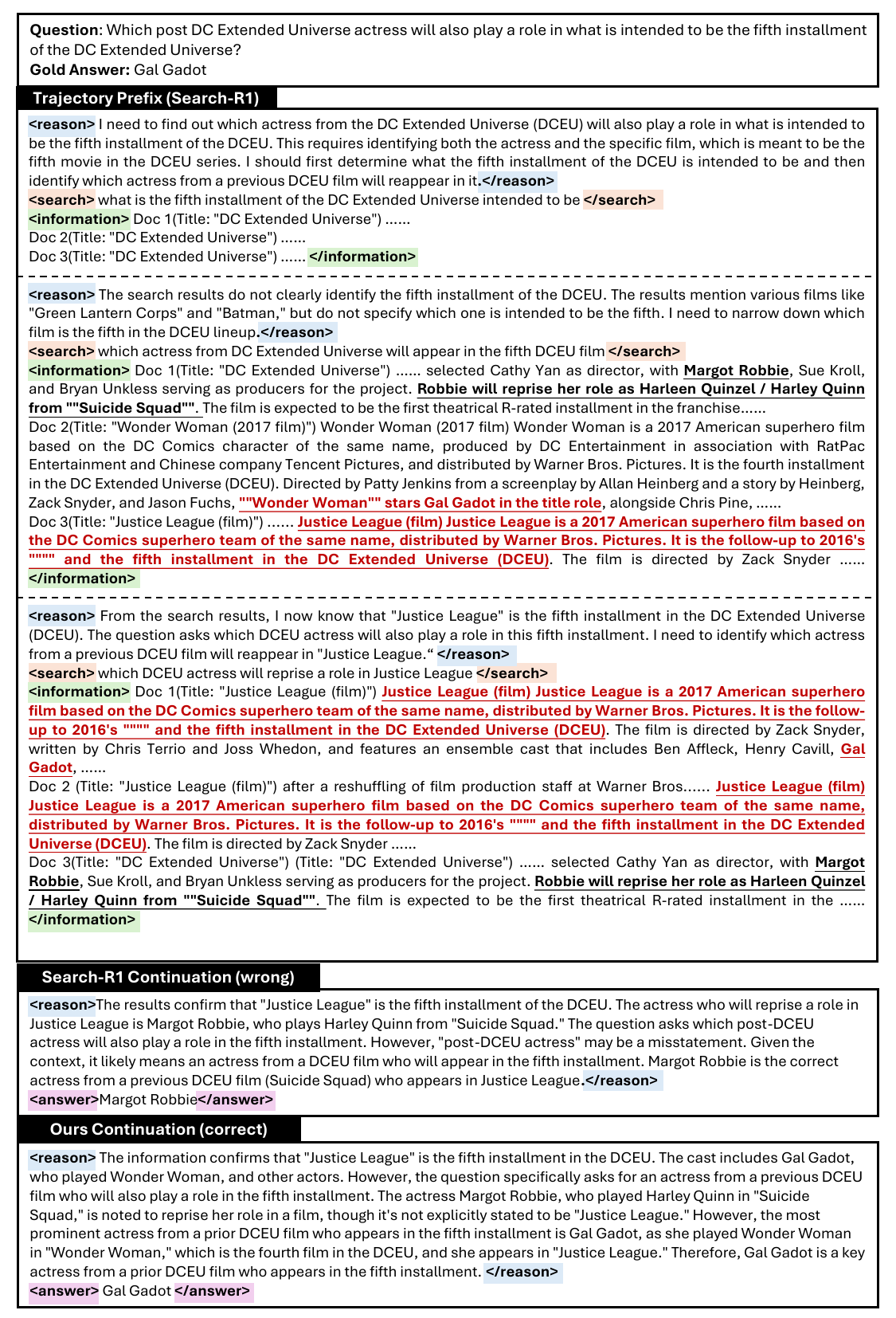}
\caption{Case study 1 where \textbf{TreePS-RAG} correctly answers the question while \textbf{Search-R1} fails, given the same reasoning prefix. In the retrieved documents, evidence supporting the correct answer is highlighted with \underline{\textcolor{red}{\textbf{red bold underlines}} }, whereas the information related to Search-R1's erroneous answer is marked with \underline{\textbf{black bold underlines}}.}

\label{fig: case-study-1}
\end{figure*}

\begin{figure*}[ht]
\centering
\includegraphics[width=0.95\textwidth]{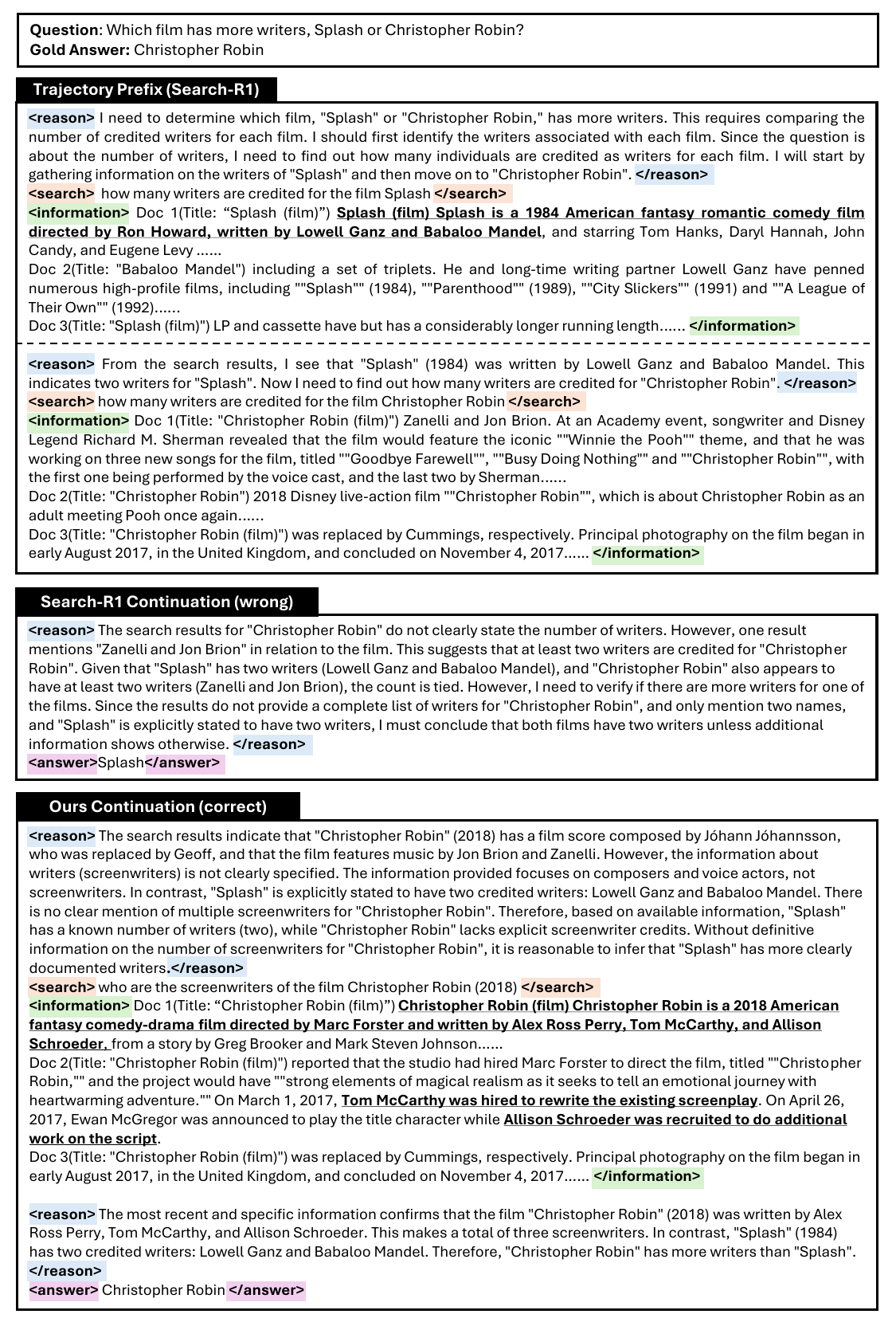}
\caption{Case study 2 where \textbf{TreePS-RAG} correctly answers the question while \textbf{Search-R1} fails to get needed information given the same reasoning prefix. In the retrieved documents, evidence supporting the correct answer is highlighted with \underline{\textbf{black bold underlines}}.}

\label{fig: case-study-2}
\end{figure*}

\paragraph{Case-Study-1}
Figure~\ref{fig: case-study-1} presents a case study for the question ``\textit{Which post–DC Extended Universe actress will also play a role in what is intended to be the fifth installment of the DC Extended Universe?}'' Answering this question requires resolving three distinct pieces of information:
(1) identifying which film constitutes the fifth installment of the DC Extended Universe;
(2) determining which actresses appear in that installment; and
(3) verifying that the actress has previously appeared in a DC film.
Under the given reasoning prefix, the model has already established that \textit{Justice League} is the fifth installment of the DCEU, but the retrieved passages contain both relevant evidence and misleading candidate mentions. When continuing from this identical prefix, \textbf{Search-R1 incorrectly grounds the answer to a spurious entity}, selecting \textit{Margot Robbie} based on a loosely related passage that mentions her role in \textit{Suicide Squad}, despite the absence of explicit evidence linking her to \textit{Justice League}.

In contrast, \textbf{our approach correctly grounds the answer in the retrieved evidence by aligning entity mentions with all required constraints.} Specifically, it identifies that \textit{Gal Gadot} previously starred in \textit{Wonder Woman} and is explicitly listed as part of the cast of \textit{Justice League}. By consistently grounding the final answer in passages that satisfy all three conditions, our method successfully produces the correct answer. This example highlights our model’s improved ability to \textbf{locate, verify, and ground answers from retrieved documents even under imperfect reasoning prefixes}, avoiding distracting entity associations that lead to incorrect answers.

\paragraph{Case-Study-2}
Figure~\ref{fig: case-study-2} illustrates another continuation-based case study for the question ``\textit{Which film has more writers, Splash or Christopher Robin?}'' Answering this question requires correctly identifying and comparing the number of credited screenwriters for both films. Under the given reasoning prefix, the model has already established that \textit{Splash} was written by two screenwriters: \textit{Lowell Ganz and Babaloo Mandel}, but the subsequent search results for \textit{Christopher Robin} fail to provide explicit screenwriter information. When continuing from this identical prefix, \textbf{Search-R1 prematurely commits to an answer despite insufficient evidence}, inferring the number of writers for Christopher Robin from loosely related information (e.g., composers and musicians) and terminating the trajectory with an incorrect answer. This behavior reflects a failure to verify missing critical information and a tendency to propagate uncertainty into the final decision. In contrast, \textbf{our approach explicitly recognizes the incompleteness of the retrieved evidence and initiates an additional verification step via query rewriting.} By reformulating the search query to directly target the screenwriter credits of \textit{Christopher Robin (2018)}, our method successfully retrieves the missing information, confirming that the film was written by \textit{Alex Ross Perry, Tom McCarthy, and Allison Schroeder}. Integrating this newly acquired evidence, our model correctly concludes that \textit{Christopher Robin} has more writers than Splash. This example highlights our method’s ability to \textbf{reflect on intermediate reasoning states, detect unresolved informational gaps, and recover from incomplete trajectories}, rather than prematurely terminating with an answer.

\end{document}